\documentclass[acmlarge]{acmart}
\renewcommand\footnotetextcopyrightpermission[1]{} % removes footnote with conference information in first column
\usepackage[utf8]{inputenc}
\usepackage[T1]{fontenc}
\usepackage{graphicx}
\usepackage{xcolor, color, soul}
\usepackage[ruled,vlined,linesnumbered]{algorithm2e}
\sethlcolor{red}
\usepackage{pgfplots}
\usepackage{subcaption}

\AtBeginDocument{%
  }

\begin{document}

\title{Hyperparameter Optimization for Federated Learning with Step-wise Adaptive Mechanisms}

\author{Yasaman Saadati}
\email{ysaad003@fiu.edu}
%%\orcid{1234-5678-9012}
\author{M. Hadi Amini*}
\email{amini@cs.fiu.edu}
\affiliation{%
  \institution{Sustainability, Optimization, and Learning for InterDependent networks laboratory (solid lab), Florida International University}
  \city{Miami}
  \state{Florida}
  \country{USA}
  \postcode{33199}
}

\renewcommand{\shortauthors}{Saadati, Amini.}

\renewcommand{\shortauthors}{Saadati and  Amini.}
\renewcommand{\shorttitle}{Hyperparameter Optimization for Federated Learning with Step-wise Adaptive Mechanisms}

\begin{abstract}
Federated Learning (FL) is a decentralized learning approach that protects sensitive information by utilizing local model parameters rather than sharing clients' raw datasets. While this privacy-preserving method is widely employed across various applications, it still requires significant development and optimization. Automated Machine Learning (Auto-ML) has been adapted for reducing the need for manual adjustments. Previous studies have explored the integration of AutoML with different FL algorithms to evaluate their effectiveness in enhancing FL settings. However, Automated FL (Auto-FL) faces additional challenges due to the involvement of a large cohort of clients and global training rounds between clients and the server, rendering the tuning process time-consuming and nearly impossible on resource-constrained edge devices (e.g., IoT devices). This paper investigates the deployment and integration of two lightweight Hyper-Parameter Optimization (HPO) tools, Raytune and Optuna, within the context of FL settings. A step-wise feedback mechanism has also been designed to accelerate the hyper-parameter tuning process and coordinate AutoML toolkits with the FL server. To this end, both local and global feedback mechanisms are integrated to limit the search space and expedite the HPO process. Further, a novel client selection technique is introduced to mitigate the straggler effect in Auto-FL. The selected hyper-parameter tuning tools are evaluated using two benchmark datasets, FEMNIST, and CIFAR10. Further, the paper discusses the essential properties of successful HPO tools, the integration mechanism with the FL pipeline, and the challenges posed by the distributed and heterogeneous nature of FL environments.
\end{abstract}

\ccsdesc[500]{Computing methodologies~Neural networks}
\ccsdesc[300]{Computing methodologies~Learning settings}
\ccsdesc[500]{Security and privacy~Privacy-preserving protocols}

\keywords{Auto-ML, Federated Learning, Hyperparameter optimization, Federated Deep Learning, Distributed optimization}

\maketitle
\thispagestyle{empty}

\section{Introduction:}
\begin{figure}[!h]
        \begin{center}
        \includegraphics[width=0.90\linewidth]{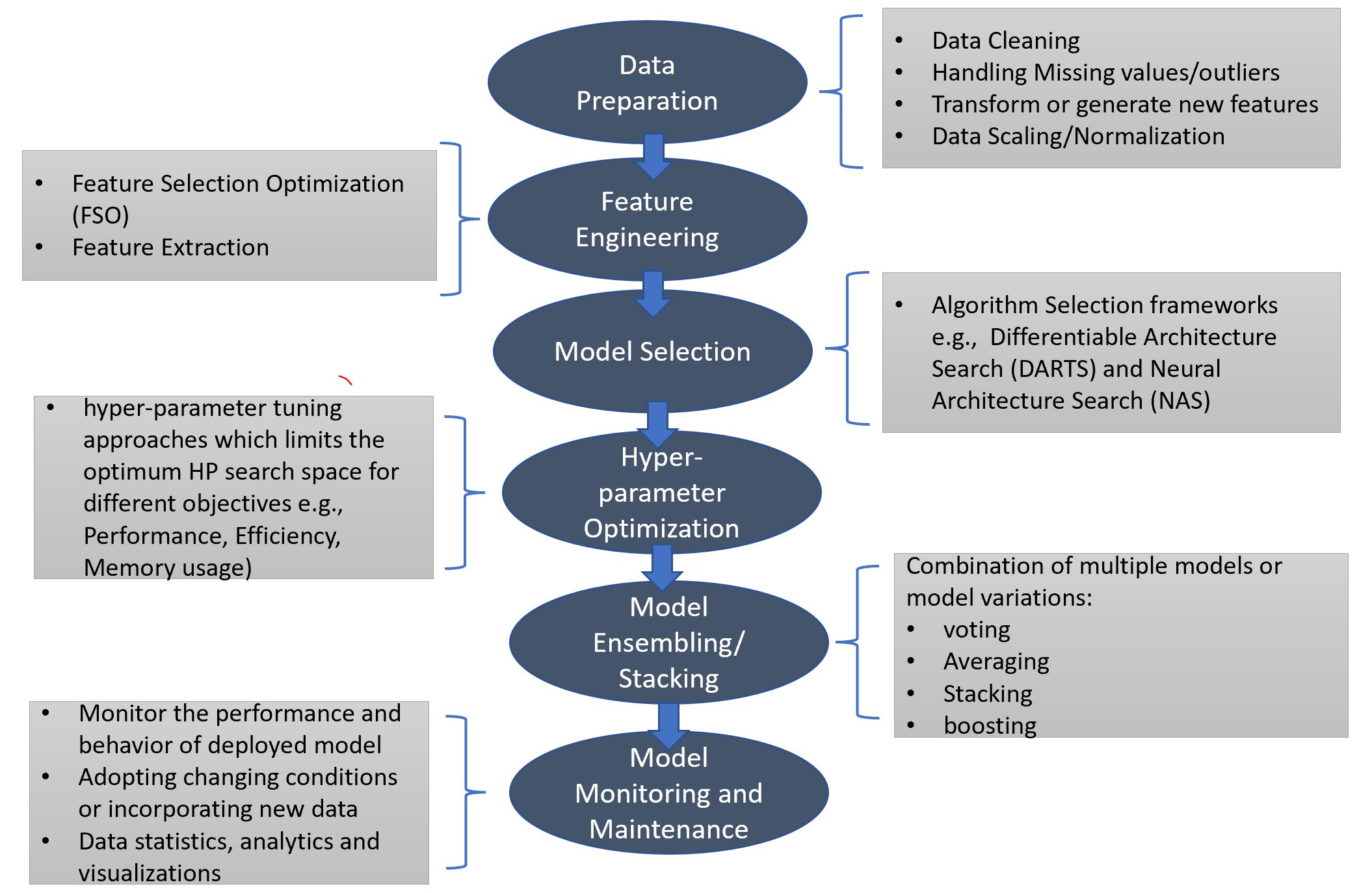}
    \caption{Different stages of Auto-ML application.}
     \vspace*{-0.41cm}
    \label{fig:Auto-ML}  
    \end{center}
        \end{figure}

\begin{figure}[!h]
        \begin{center}
        \includegraphics[width=0.5\linewidth]{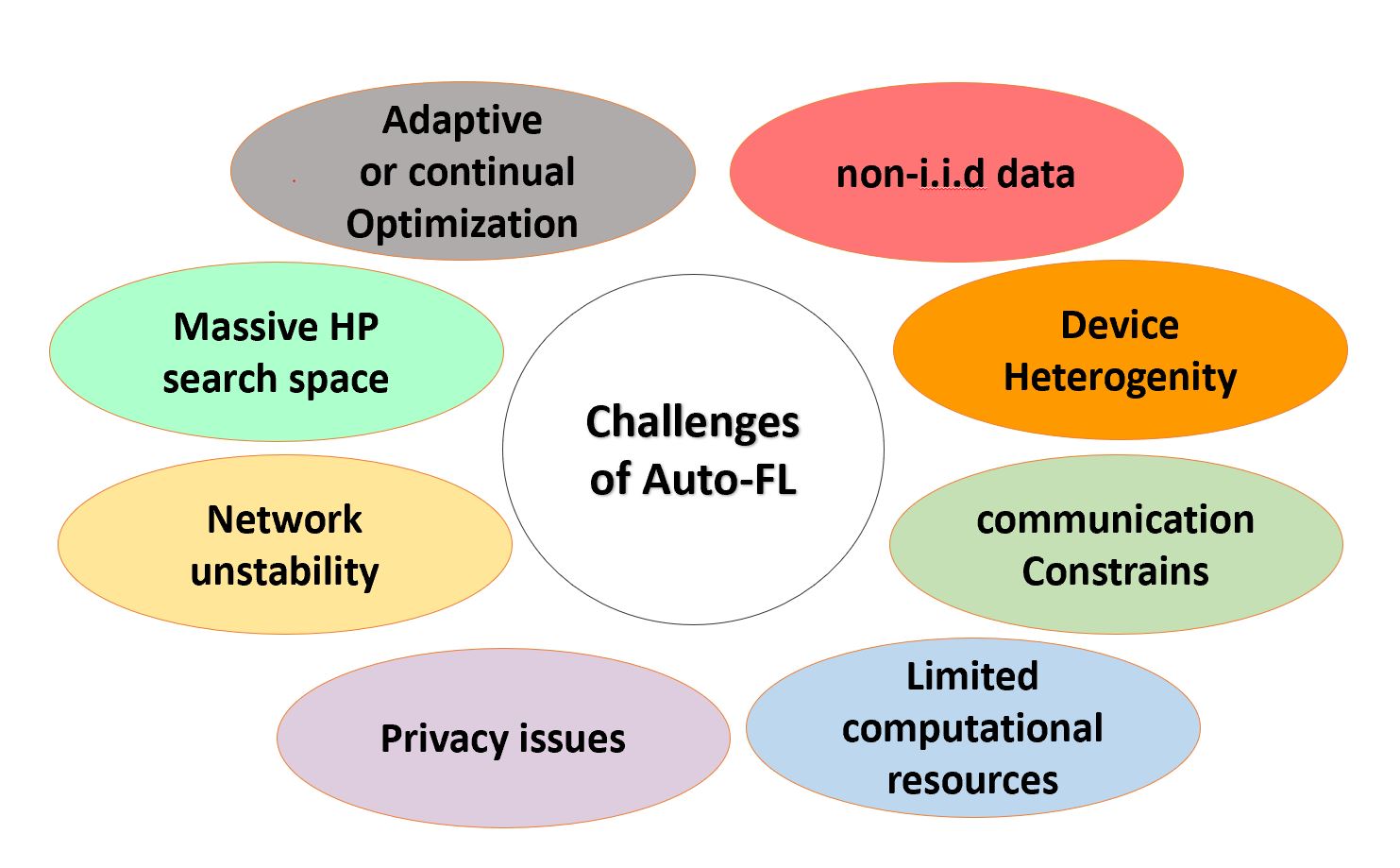}
    \caption{Challenges of Automating the FL setting.}
     \vspace*{-0.41cm}
    \label{fig:Auto-FL_challanges}  
    \end{center}
        \end{figure}

The rapid data production from diverse sources, such as edge IoT devices and large organizations, presents new challenges for centralized data aggregation. Centralized approaches are often inefficient due to constraints in computational resources and communication bandwidth. Moreover, privacy concerns and the potential for data leakage deter many organizations and individuals from sharing their datasets. Federated Learning (FL) addresses these issues by enabling secure, decentralized model training directly on edge devices. In FL, a global server first distributes the initial model parameters to each participating device. Each device is updating the global model by local training on each client's raw data. Next, the local model parameters are sent back to the global server for aggregation and to update the global model. The global model’s parameters are then shared with the edge devices to refine their local models \cite{5} \cite{6}. This iterative process continues until predefined stopping criteria are met.  By keeping the data local, FL preserves user privacy and reduces the need for transferring sensitive information. Additionally, FL facilitates the creation of more personalized models, which can be beneficial in various applications, including GBoard\footnote{Google Keyboard}, healthcare, and FinTech where customized models are required for each user. Despite its promise of enabling distributed learning, FL still faces several unresolved challenges that require further research \cite{5}. \\
To address the challenges of FL, the integration of classical ML optimization techniques in FL environments has gained significant attention. \textbf{Automated Machine Learning (AutoML)} aims to reduce the need for manual intervention by automating various stages of the ML pipeline, including data preparation, model selection, and hyper-parameter optimization (HPO) \cite{2}. The different stages of AutoML are depicted in Figure \ref{fig:Auto-ML}. AutoML accelerates and optimizes the overall ML development cycle, helping both experts and non-experts by reducing repetitive tasks, shortening training time, enhancing model accuracy, and optimizing resource usage \cite{zoller2021benchmark,7}. \\
\textbf{Automated Federated Learning (Auto-FL)} extends the principles of AutoML to FL, aiming to minimize manual tuning in federated settings. However, Auto-FL faces unique challenges, primarily due to the non-Independent and Identically Distributed (non-IID) nature of the data, resource constraints, and the high number of parameters in FL systems. Moreover, data heterogeneity which stems from variations in data distribution across clients, complicates key tasks such as model selection \cite{11} and hyperparameter optimization (HPO) \cite{9}. Unlike centralized ML, the non-IID characteristic in FL makes it difficult for Auto-FL algorithms to generalize across clients, as each client’s data can differ significantly.
Additionally, deploying resource-intensive AutoML algorithms in FL environments is often infeasible due to the limited computational power and memory capacity of edge devices (e.g., IoT devices, mobile phones) \cite{5}. The heterogeneity of clients' resources, combined with communication constraints, further complicates the optimization process. In FL, the search space for hyper-parameters expands significantly due to the inclusion of both local and global parameters, leading to a prohibitively high search time. On the other hand, traditional Auto-ML approaches that rely on proxy data or extensive information exchange between clients are not suitable for FL due to privacy concerns and the strict requirement to keep data localized. This severely restricts the scope of optimization techniques that can be applied in FL. The challenges associated with Auto-FL are summarized in Figure \ref{fig:Auto-FL_challanges}.

HPO is a core component of AutoML and plays a critical role in improving model performance and generalization. It has been successfully applied to optimize the hyperparameter (HP) sets of various ML models and deep neural networks (DNNs) \citeN{2,3}. HPO algorithms aim to reduce the hyperparameter search space by eliminating insignificant configurations (also referred to as non-optimal hyperparameters). This reduction in search space leads to decreased search time and increased accuracy. Consequently, identifying a near-optimal set of hyperparameters has a significant impact on the overall performance of FL \cite{21}. 

FL HPs can be categorized into three main groups \footnote{A full list of common HPs used in FL systems can be seen in Table \ref{table2}}: \\
\textbf{1-	Local HPs:} These HPs are related to the general models in local nodes, such as Learning Rate (LR) \cite{20}, Batch Size, momentum, weight decay, dropout, and the regularization term for DL. \\
\textbf{2-	Global HPs:} HPs associated with the FL paradigm, related to the global server or aggregation, e.g., the total number of clients, cohort size (number of clients per round), communication rounds, and local steps per round. \\ 
\textbf{3-	Communication management HPs:} HPs that manage the connection between local and global HPs, such as lambda $\lambda$ (degree of personalization), security protocols, and communication type. 

HPs involved in the FL framework can be optimized based on multiple objectives, depending on the selected set of HPs. Applying conventional Auto-ML methods, which evaluate nearly all combinations of HPs, is impractical in FL environments. Moreover, due to the decentralized nature of training and validation in FL, collecting feedback from each client regarding the chosen HPs is time-consuming, computationally heavy, and communication-inefficient, potentially leading to convergence issues \cite{17,19,20}. Although many existing HPO approaches require all nodes to participate in every communication round, this is not feasible in FL due to resource-limited devices and network unavailability \cite{21}. Communication overhead between clients and the server is another issue in FL, and this problem amplifies exponentially when testing different HP configurations for HPO. \\
It is important to note that integrating Auto-ML with FL increases complexity. The FL framework involves a large cohort of participating nodes, each with a substantial number of HPs. Identifying a near-optimal set of HPs for a specific optimization goal amidst this extensive array of HP combinations represents a major challenge in Auto-FL that needs to be addressed \citeN{3,5,khodak2021federated}. 

The primary focus of this paper is to address the deployment of HPO toolkits on FL platforms. While it is common to exploit and compare HPO toolkits as baselines for tuning HPs in classic ML approaches, to the best of our knowledge, prior works have not specifically deployed available HPO tools in the FL setting. Despite the significance of automating FL platforms, particularly in HPO, there are few publications in this area, with most focusing on specific objectives or applications \cite{8,9,17,19,20,21}. On the other hand, although comprehensive baseline comparisons exist for different HPO tools in ML \cite{7,3}, no prior work has been found on integrating them into FL systems. To address this challenge, this paper integrates and optimizes two major HPO tools, ``Raytune'' and ``Optuna'', for FL. Furthermore, the paper discusses the properties and requirements for selecting suitable HPO tools for FL and presents novel ideas and methods for applying and integrating these toolkits. It is worth mentioning that while the FL optimization algorithm employed in this paper is ``FedAvg,'' the ideas presented can be extended to other optimization algorithms in FL with a trusted server, such as ``FedProx'' \cite{li2020federated}, and ``FedDP'' \cite{wang2020tackling}. 

The novelty of this study lies in the comprehensive comparative analysis of $Optuna$ and $Ray Tune$ in the FL context. We assessed their impact on overall model performance and examined their ability to find near-optimum HPs, using clients' feedback in each communication round, along with global model feedback to accelerate the HP search process. Moreover, a step-wise adaptive mechanism has been introduced that could update the HPs according to the feedback received at each local SGD step.   The findings not only shed light on the advantages and limitations of these HPO tools but also provide insights into their applicability in FL settings. The main contributions of this paper are listed as follows:
\begin{itemize}
\item It provides a practical understanding of deploying HPO tools for FL libraries and assists in selecting the appropriate tool based on the requirements and limitations of the FL system.
\item It advances the field of FL research by highlighting the benefits and drawbacks of HPO in enhancing the efficiency of FL models. This is achieved by comparing the functionality and integrating two major HPO tools for the FL platform. To this end, we evaluated our HPO-FL framework on two non-i.i.d benchmark datasets (FEMINIST and CIFAR10).
\item We have designed a step-wise feedback mechanism for adaptive HPO in the FL environment to expedite the HP tuning process. In this approach, the HPO tool can update the HPs based on local feedback, in addition to the global training and validation feedback obtained from all clients participating in the current communication round. 
\item Also, we've introduced a novel client selection strategy in the Auto-FL framework which has significantly improved the overall running time. 
\item To limit the search space, we have employed lightweight strategies, such as low-fidelity mode, to address the curse of dimensionality in the HPO-FL problem. Low-fidelity mode narrows down the HP search space by clustering HP configurations in each training round.  
\end{itemize}

In the next section, we provide a brief literature review on the various stages in which Auto-ML can be applied in the FL setting. We explain our proposed framework, methodology, and its effectiveness in Section \ref{3}. Further, we explore different available Auto-ML tools and investigate their suitability for adoption in FL. The experimental setup of our Auto-FL framework, dataset description, list of existing hyperparameters in FL, and implementation details are presented in Section \ref{4}. Moving forward, we report the results in Section \ref{5}. Section \ref{6} concludes the paper and discusses future directions.

\section{Related Work} \label{2}
Although Auto-ML and Automated Deep Learning (Auto-DL) algorithms are successful tools and several surveys have explored and compared them for classic ML and Deep Learning (DL) problems \cite{7,3}, there is less focus in the literature on Auto-FL \cite{12}. The majority of prior works studying Auto-FL can be classified into the following categories: 1- Feature Selection Optimization, 2- HPO, 3- Neural Architecture Search (NAS), and 4- Automated Aggregation. While a few studies have developed Auto-ML for FL in various stages, the main focus of this paper is on integrating HPO into the FL platform. We briefly review prior works that have applied and developed HPO for the FL setting.

\subsection{Hyper-parameter Optimization (HPO)}
In \cite{khodak2021federated}, authors formulated the HPO in FL, investigated the challenges, and introduced the baselines for Auto-FL \cite{khodak2021federated}. They deployed the weight-sharing approach from the NAS for addressing the HP search in FL-HPO problem. The proposed method called $FedEx$ has optimized the Learning Rate (LR) of local devices in personalized FL(PFL). The shortcomings of $FedEX$ approach are as follows: 1- it is only adaptable in FL algorithms with separate local training and aggregation such as $FedAvg$, and 2- it may not be effectively applied for global HPs (server-based) and communication management HPs (such as PFL). In \cite{saini2021fms}, authors have exploited Auto-ML for model selection and HPO of FL framework in healthcare systems. The authors claimed that the customized local model selection leads to a more personalized recommendation system. They only applied Auto-ML to the server with $AzureML$ Auto-ML toolkit. Moreover, the Authors in \cite{nilsson2018performance} have used Random Search(RS) as an HPO method for tuning different HPs in FL optimization algorithms, e.g.,  $FedAvg$, $CO-OP$ for MNIST dataset with a Non-i.i.d distribution. Although the article reported successful results in finding the near-optimum HPs(batch size, learning Rate (LR), learning rate decay, and the number of local epochs), it is time-consuming and requires 1200 communication rounds to evaluate each of these HPs.      
 Another study that has applied HPO to a single-shot FL \footnote{FL system with just one communication round between central server and local clients} is FLoRA \footnote{Federated Loss Surface Aggregation} \cite{21}. This research has investigated different ML models (non-neural network models such as  Multi-Layer Perceptron (MLP), and SVM) for tabular data and demonstrated the performance of FL-HPO in an asynchronous FL setting.  In \cite{13}, the particle swarm optimization (PSO) algorithm has been exploited as an HPO method for FL data in autonomous vehicles’ systems. Further, \textit{Optimized Quantum-based Federated Learning (OQFL)} approach has shown its effectiveness against adversarial attacks while tested on \textit{CIFAR10} and \textit{FEMNIST} datasets. Further, authors in \cite{9} used grid search and Bayesian optimization on the non-i.i.d MNIST dataset in an industrial FL setting. This study also compared both local and global HPO and asserted that global HP optimization would slightly exceed local HP optimization in terms of test accuracy but it is more efficient for the communication cost. Nevertheless, this hypothesis requires more exploration as it has been evaluated on one dataset.\\ 
Some studies under the category of PFL have focused on optimizing communication management HPs especially $\lambda$ which regulates the connection between global and local models. $\lambda$ can be used to adjust the personalization level and contribution of local and global HPs in shaping the FL model. In \cite{14}, a PFL optimizer, referred to as ``Ditto'', is developed to maximize the robustness against training-time attacks and increase fairness to enhance the performance across device heterogeneity as well as accuracy. Further in \cite{15}, a novel bi-level approach has been proposed which divided the global and local optimization for the aim of addressing data heterogeneity in Personalized FL setting. In this approach, global model optimization has not been altered while the local model optimization is enhanced according to its corresponding data distribution. The results illustrated that this approach called ``pFedMe''  has achieved higher accuracy with faster convergence speed in comparison with. The goal of the experiments was to tune the following HPs: $R$ (local computation round), $K$ (computation complexity), $|D|$ (mini-batch size), and $\lambda$ HPs. The Authors in \cite{8} have applied local HPO to cluster similar clients for the electricity load forecasting problem. The results show that the nodes with similar optimal HPs (such as the learning rate and the number of local epochs) could have similar historical data and also HP-based clustering could lead to a faster convergence in the FL setting. Authors in \cite{zhang2021automatic} have explored the HPO-FL problem from a system perspective. In the first phase, they tried to formulate and measure the run time, communication overhead, and computational complexity with different HP configurations e.g., the number of participants in each round and model structure. In the second phase, they introduced FedTuning, an HPO algorithm for FL, which could consider the three mentioned metrics while searching for the near-optimum HP set. This paper hasn't considered a case in which the participants are heterogeneous and could show a different behavior. Further, the authors in \cite{agrawal2021genetic} addressed the device heterogeneity problem by clustering nodes (and their associated HPs) into different clusters. They have introduced their algorithm called Gentic CFL which uses the Genetic algorithm as an HPO tool for each cluster, separately.  

\subsection{Neural Architecture Search (NAS)}
Since the neural structure directly impacts the overall performance of DNNs, NAS is one of the most notable aspect of Auto-ML \cite{13}. NAS aims to discover a near-optimal DNN structure and its corresponding HPs with minimal manual intervention. While it is common to leverage existing DNN models for FL environments, the non-i.i.d distribution of FL datasets changes the DNN architecture used in FL framework as compared with traditional DNNs. Therefore, numerous model structures with diverse HPs should be explored in FL setting. However, this process is time-consuming, resource-intensive, and computationally expensive. Consequently, automating both the model design and HPO procedures can potentially lead to higher precision, faster convergence, and/or increased robustness in FL systems \cite{11}. It is worth noting that NAS has only been developed for DNNs. Hence, It may not be applicable to other ML models such as SVM or random forest.  There are three common strategies for applying NAS to the DNNs: 1- Reinforcement learning (RL)-based NAS 2- Gradient Descent (GD)-based NAS, and 3- Evolutionary (EA)-based NAS. Prior works have applied NAS for FL settings \citeN{10,11,seng2022hanf}. Authors in \cite{10} and \cite{11}  proposed a GD-based method because of its rapid search ability for finding the optimum HPs. The results verified the superiority of NAS compared to pre-defined FL architectures, even in the early rounds of training. In \cite{seng2022hanf} a joint framework has been designed that contains both HPO and NAS components, referred to as \textit{HANF}. It deploys the $n$-armed bandit approach as an HPO method for both searching and evaluation stages \cite{seng2022hanf}.

\begin{figure}[]
  \includegraphics[width=\textwidth]{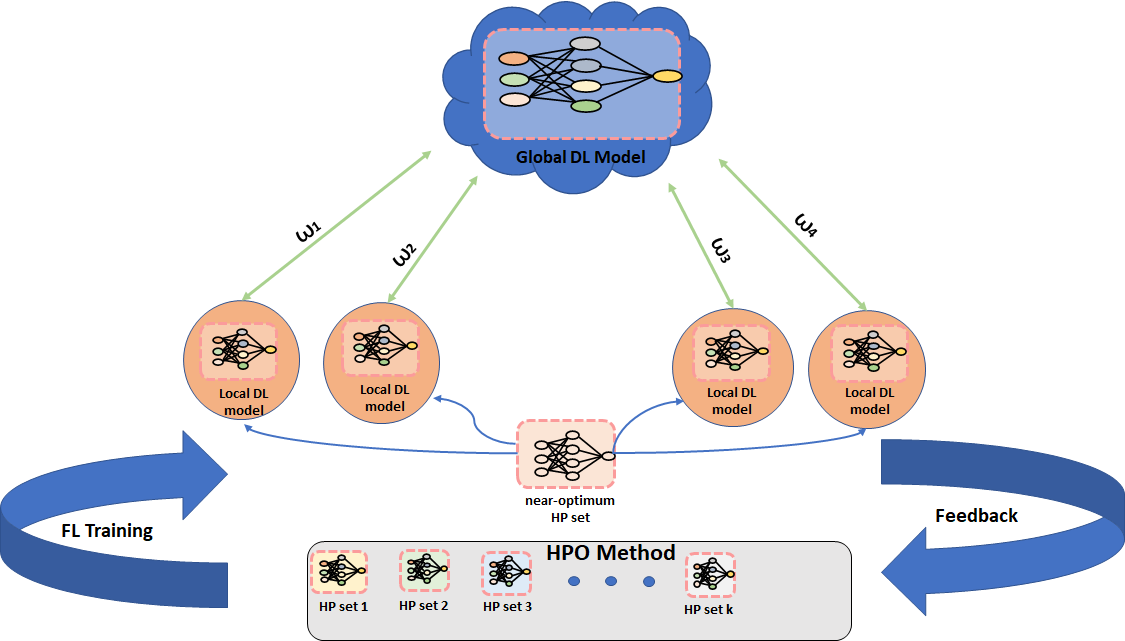}
  \caption{Overall Framework of Applying HPO in FL system. All local models used in this study and the majority of literature, are trained on the same DNN architecture but with different sets of HPs for evaluation. The HPO toolkit is inside the global trusted server.}
  \label{figure:overview}
\end{figure}
\vspace{-3mm}

\section{Methodology} \label{3}
\subsection{Formulation of the Proposed HPO-FL Method}
HPO aims to find the HP set $\mathbf{h}^*$ to optimize the FL objective function, $f$. Equation \ref{1_formula} shows the HPO formulation for FL problem: \\
\begin{equation} \label{1_formula}
\mathbf{h}^* = \arg\min_{\mathbf{h} \in \mathcal{H}} \sum_{c \in \mathcal{C}} \frac{|\mathcal{D}_c|}{|\mathcal{D}|} f(\mathbf{h}, \mathbf{w}^*, \mathbf{X}_c, \mathbf{y}_c),
\end{equation}
 \noindent where $h^*$ is the near-optimum HP set and $C$ represents the set of all clients, and  $\mathcal{D}_c$ shows the raw data of the local client $c$. Further, $\mathbf{X}_c$, $\mathbf{y}_c$, and $\mathbf{w}^*$ represent the training set, test set, and optimal weights respectively. The loss will be calculated according to the local models and the data distribution of each client.  $h$ is the list of possible HP sets that will be evaluated. The main goal of HPO-FL is to find the $h^*$ within a limited search space while minimizing the search time.  
\subsection{Overview of the Proposed Framework and Deployed Tools}
The main focus of this paper is applying HPO in a DFL setting to maximize overall accuracy. Most of the prior works in the FL-HPO merely concentrated on introducing self-designed HPO algorithms that only adapt to their proposed FL solution. To the best of our knowledge, this study is among the leading efforts to explore available open-source HPO tools (designed for classic ML/DL problems), in the FL setting. While there are some studies focused on applying these HPO tools for many ML/DL toy problems \cite{2}, there is a gap in the literature for investigating the effectiveness of these practical Auto-ML/Auto-DL toolkits in FL setting. The importance of integrating generic Auto-ML tools with FL is not only due to the adaptability of developed HPO toolkits but also because of their fast-paced, easy-to-deploy, and up-to-date HPO for multiple ML benchmarks. \\
Figure \ref{figure:overview} presents an overview of our proposed method for integrating generic HPO toolkits in the FL setting. The HPO module (gray box) generates an initial set of hyperparameters (HPs) to be evaluated by the FL system. The FL system then updates the local models’ HPs based on the output of the HPO tool. Local models are trained on the client side, and the resulting parameters are transferred to the global server. Additionally, each client sends feedback to the HPO toolkit, which is assumed to reside on a trusted server. This feedback includes the training and testing loss calculated during the most recent local training round. \\
Although the first communication round does not yield final validation results, the interim feedback can be valuable for refining the chosen HPs during the optimization process. This step-wise local feedback mechanism is applied for each individual HP. For example, if the goal is to optimize accuracy based on learning rate and weight decay, each client sends three additional feedbacks to the HPO toolkit: one for the current HP combination, and two for the next best choice of each HP, while holding the other constant. This mechanism narrows the HP search space and accelerates the HPO process in the early stages of the search. 
It should be noted that while the HPO toolkit could store all feedback history for future HP optimization, only the most recent performance feedback is used for updating HP suggestions in this approach. This decision aligns with the findings in \cite{khodak2021federated}, where the effectiveness of using the last performance feedback, compared to aggregating a weighted sum or average of all previous feedback, was evaluated. The results demonstrated that maintaining an extensive feedback history did not significantly improve overall performance. As a result, here we adopted a step-wise approach to local HP updates, as opposed to conventional HPO algorithms like Random Search, which only apply global updates. \\
After collecting step-wise feedback from the clients, the average of all local feedback is stored in the HPO tool. This is necessary because, at the end of each communication round, the average of all local model weights is computed and evaluated. \footnote{In this study, $FedAvg$ is used as the aggregation method.} The global server asynchronously aggregates the updated weights, averages the received local weights, and sends the updated parameters back to each client. At this point, the calculated global loss for both the training and test datasets is sent back to the HPO tool as the second feedback. Once the first communication round is completed, the next round follows the same procedure. This iterative process continues until the stopping criteria are satisfied.

\subsection{Client selection in Auto-FL }
Due to device heterogeneity, Some devices might be faster due to their computation and communication superiority. For handling Device heterogeneity which could be in terms of memory, connection bandwidth, and computational power, Asynchronous FL has been picked to avoid the Straggler \footnote{Straggler nodes are the clients that stall the whole FL paradigm due to their slow performance.} effect \cite{5}. In this work, the HPO toolkit updates the feedback set for an HP set as the slower nodes send out their parameters to the central server. Since most HPO toolkits require one feedback for each set of HPs, The challenge is to handle the parallel continuous feedback. We have addressed this problem by 1- grouping the clients dynamically based on their training time and 2- the step-wise adaptive HPO mechanism which allows multiple performance evaluations simultaneously for different HP configurations. This means that whenever the HPO toolkit gets feedback for an HP subset $h^*_1$ from clients' group $g_1$, The HPO toolkit will send a new set of HPs $h^*_2$ according to the previous feedback to $g_1$. So the group $g_1$ participants will update their HPs as well as their parameter weights which have been calculated by the central server. When the group $g_2$ is prepared to send their feedback for $h^*_1$, the HPO toolkit will update its prior feedback for $h^*_1$ by averaging it among all the previous feedbacks received. This way,  the HPO process is sped up in the FL setting while it could benefit from the parallel evaluation of several HP sets locally. \\
%\textcolor{red}{here}
Another important point to mention is the second feedback the HPO toolkit should receive at the end of each communication round. Plus to the local feedbacks $lf_j$, the global feedback $gf_j$  is essential. It is true to say, that global feedback should be taken into account more than local feedbacks as it plays a more significant role in the overall performance of FL. Consequently, We have used a weighted averaging method for calculating the overall feedback in communication round j. In this method, the weight of global feedback will be calculated based on the number of participating nodes in the corresponding group $g_j$ in round j. Algorithm \ref{HPO_FL} demonstrates the overview of HPO-FL pipeline and the model selection procedure. In the end, the output is a set of HPs $h^*$ and weights $w^*$ which optimizes the $f$ function in the equation \ref{1_formula}. 

As represented in the overview of this work, In the following section we’ll explore a variety of popular HPO toolkits which have a good reputation for finding the near-optimum solution for classic ML/DL problems and have been exploited for many benchmarks.  Next, The possibility of deploying them as a potential Auto-FL tool has been discussed. \\

\begin{algorithm}[H]
\label{HPO_FL}
\caption{HPO-FL pipeline}
\SetAlgoLined
\KwIn{An HPO toolkit $H$ Set of clients $C$,  training set $x_c$ , validation set $v$ and Client weight $w_c$ $(\forall c\in) C$}
\KwOut{Trained model and the optimized HP set ($w^*, h^*$)}
Initialize weights $w_j \leftarrow w_0$\;
Initialize global feedbacks                 $gf_j \leftarrow w_0$\;
Initialize local feedbacks          $lf_j \leftarrow w_0$\;
$h_j \leftarrow \text{first set of HPs from } H$\;
\For{$ \text{communication round } j \text{ from } 1:K$ }{
    $w_c \leftarrow w_j$\;
    \For{$c \in C$}{
        $c^h_j\leftarrow \text{client\_train\_steps}(h_j, w_c, x_c)$\;
        $\text{update } g_j; \text{   //   check for the clients that are ready} $\
    }
    \For{$c \in g_{1:j}$}{
        $lf_j \leftarrow \frac{1}{c_j}\sum_{c=1}^{c_j}(l([c^h_j]) + lf_j$\;
        $w_j \leftarrow \frac{\sum_{W_c \in C_j}}{|C_j|} $\; }
    $gf_j \leftarrow l(w_j,v) + gf_j$\; 
    $h_j \leftarrow  H(lf_j,gf_j,h_j); \text{ // H update the HP set based on the local and global feedback}$\
 }

\end{algorithm}

\vspace*{0.5cm}

\section{Experimental Setup}\label{4}
\subsection{Auto-ML tools}
There are dozens of FL packages introduced by different organizations every year. We briefly present some of the most popular FL open-source libraries in Table \ref{table2}. Conversely, there are several well-known tools specifically designed for Auto-ML and/or Auto-DL. Table \ref{table1} depicts a comprehensive comparison of Auto-ML toolkits. The libraries discussed here include TPOT, GAMA, Auto-SKLearn, AutoGluon, FLAML, Auto-Keras, Auto-PyTorch, H2O, and finally, Optuna and RayTune. The strengths and limitations of these libraries, and insights into why some of these tools may or may not be suitable candidates for FL environments, have been discussed in \ref{table1} as well.

\begin{itemize}
    \item \textbf{TPOT} is an Auto-ML library that uses a tree-based structure and a Genetic Programming stochastic global search. It is designed for both classification and regression tasks and supports both Feature Selection and Model Selection. TPOT-NN is an extension of TPOT, which works for neural networks. However, it has not been thoroughly tested on DL tasks.
    
    \item \textbf{GAMA} is an AutoML tool that automates the data preprocessing step and offers HPO for some ML algorithms. It is capable of combining and creating ML algorithms in the form of Ensembles, supporting Binary Classification, Multiclass Classification, and Regression. However, GAMA is not designed for DNN structures, and consequently, it is not suited for DL. It is limited to tabular datasets.
    
    \item \textbf{Auto-SKLearn} is built on top of scikit-learn and is capable of identifying learning algorithms for new datasets, as well as conducting HPO. It includes 15 classification algorithms, 14 feature preprocessing algorithms, and utilizes advanced approaches like Bayesian optimization, meta-learning, and ensemble learning. However, Auto-SKLearn does not support Auto-Deep Learning (Auto-DL) or neural networks. It only operates on Linux systems and has relatively high computational complexity.

    \item \textbf{AutoGluon} is an AutoML tool built in PyTorch, primarily aimed at automated stack ensemble, deep learning, and real-world applications involving image, text, and tabular data. It is user-friendly for beginners and highly extendable for experts. It offers the ability to tune the number of hidden layers, activation functions, dropout rate, and batch normalization in DL models. AutoGluon also supports HPO, model selection/Ensemble, Neural Architecture Search (NAS), and relatively low computational complexity data processing. AutoGluon is a strong candidate for multi-class classification and regression in DL, although to our knowledge, it has not yet been tested in distributed ML settings.
    
    \item \textbf{FLAML}, introduced by Microsoft, offers HPO and model selection options for both general ML/DL structures. It is adaptable to heterogeneous evaluation costs and can handle complex constraints, guidance, and early stopping, allowing for personalized model modification and metric establishment. FLAML is typically low in computational cost and relatively fast in Auto-tuning. However, FLAML has not been tested in FL settings thus far.
    
    \item \textbf{Auto-Keras} is a Python library with a focus on NAS and is based on Bayesian optimization. Built on top of Keras, it is relatively fast in finding near-optimal HPs. Auto-Keras is easy to implement and is designed for non-experts. It targets both NAS and HPO for DNNs, applicable to text, images, and tabular data. However, it does not address general ML models and lacks support for cross-validation and the macro F1 validation metric. It has not yet been tested for distributed ML.
    
    \item \textbf{Auto-PyTorch} is a PyTorch library primarily focused on NAS for DL, similar to Auto-Keras. It uses multi-fidelity optimization for NAS and accelerates HP search through Bayesian optimization, meta-learning, and ensemble learning. Auto-PyTorch supports regression, binary, and multi-class classification using tabular, image, and text data. However, the available version is incomplete, lacking cross-validation and macro F1 validation metrics, and does not address general ML models.

    \item \textbf{Ray Tune} is a Python library designed to assist with distributed HPO and Auto-ML. It offers a consistent API and interface for various HP tuning algorithms such as random search, grid search, and Bayesian optimization. Beyond HP tuning, Ray Tune also provides Auto-ML algorithms for model selection and HP tuning based on input data and target tasks. Notable algorithms include Population-Based Training (PBT) and BOHB (Bayesian Optimization and HyperBand) have been offered on this toolkit. Ray Tune supports distributed training and can be integrated with various ML frameworks and algorithms such as TensorFlow, PyTorch, and XGBoost, making it versatile and user-friendly.

    \item \textbf{Optuna} is a lightweight and flexible HPO library offering several search algorithms and pruning strategies. Its simplicity and ease of use make it highly suitable for FL scenarios. Further, Optuna's support for distributed optimization is a significant advantage when dealing with a large cohort of clients participating in the FL system.
    
\end{itemize}

\begin{table}
\centering
\footnotesize
\caption{Overview of popular open-source Auto-ML toolkits}
\label{table1}
\begin{tabular}{|p{2.5cm}|p{4.5cm}|p{4.5cm}|}
\hline
\textbf{Library} & \textbf{Strengths} & \textbf{Flaws} \\
\hline
TPOT\footnote{https://github.com/trangdata/tpot-nn} & 
- Supports both Feature Selection and Model Selection - Designed for both classification and regression - Have a DNN extension \footnote{TPOT-NN} & 
- it's not specifically designed for Deep Learning.
- Haven't been structured for distributed systems. \\
\hline
GAMA &
- capable of uniting several tuned ML structures (Ensemble)
- supports Binary Classification, Multiclass Classification, and Regression &
- not designed for Deep learning
- Just works with tabular datasets.
\\
\hline
Auto-SKLearn &
- Capable of finding learning algorithms for unseen datasets
- supports Bayesian optimization, meta-learning, and ensemble learning.
- can automate the data scaling, encoding of categorical parameters, and missing values.&
- Doesn't support Auto-DL and NNs.
- Limited to Linux operating systems.
- High computational complexity. \\
\hline
AutoGluon\footnote{https://github.com/awslabs/autogluon} &
- Quick and easy to run on raw data.
- Capable of tuning DL architectures. &
- Immature (version 0.0).
- No expert knowledge required.
- Limited to FL setting. \\
\hline
FLAML\footnote{https://github.com/microsoft/FLAML} &
- Offers HPO and model selection options.
- Supports general ML and DNNs. &
- Not tested for FL.
- Complex constraints and guidance.
- Relatively low computational cost. \\
\hline
Auto-Keras\footnote{https://github.com/keras-team/autokeras} &
- Fast in finding near-optimum HPs.
- Targets NAS and HPO for DNNs. &
- Does not address general ML models.
- Does not support cross-validation and macro F1 metric.
- Applicable for text, images, and tabular data. \\
\hline
Auto-Pytorch\footnote{https://github.com/automl/Auto-PyTorch} &
- Supports binary/multi-classification and regression.
- Uses multi-fidelity optimization for NAS. &
- Does not support cross-validation and macro F1 metric.
- Applicable for text, images, and tabular data. \\
\hline
H2O &
- Exploits grid search for model selection.
- Supports DL, ML, and XGBoost models. &
- Supports Java, Python, and R languages.
- Successful in binary classification and regression.
- Supports tabular data preparation.
- Used in FL settings. \\
\hline
RayTune &
- fully support HPO for distributed ML.
- can handle large cohorts of federated data. 
- supports most ML libraries such as XGBoost and MXNET 
- includes powerful novel algorithms such as Evolutionary and HyperBand
- visualize the results with TensorBoard &
- limited documentation
- have complex setup and configurations
\\
\hline
Optuna &
- Easy to work with for both experts and beginners.
- is very lightweight. 
- the user can define the search space with Python 
- supports distributed systems and parallelism &
- Not scalable for very large distributed data
- lack of state-of-the-art search algorithms
\\
\hline
\end{tabular}
\end{table}

\begin{table}
\centering
\caption{Overview of the most popular open-source Auto-ML toolkits} 
\label{table2}
\begin{tabular}{ | l | l | l | }
\hline
	\textbf{NAME} &  \textbf{FOUNDER} & \textbf{SUPPORTING LANGUAGES} \\ \hline
	FLOWER \footnote{https://github.com/adap/flower} & Flower.dev & Pytorch/Tensorflow/Keras/MXnet/ScikitLearn \\ \hline
	PySyft \footnote{https://github.com/OpenMined/PySyft} & OpenMinded & Python3(PyTorch/Tensorflow) \\ \hline
	OpenFL \footnote{https://github.com/intel/openfl} & Intel & Python (PyTorch/Tensorflow/Keras) \\ \hline
	TensorFlow Federated \footnote{https://github.com/tensorflow/federated} & Google & Tensorflow \\ \hline
	IBM FL \footnote{https://ibmfl.mybluemix.net} & IBM & Python (PyTorch/Tensorflow/Keras) \\ \hline
	FedML \footnote{https://github.com/FedML-AI/FedML} & FedML, Inc & Python \\ \hline
	FederatedScope \footnote{https://github.com/alibaba/federatedscope} & Alibaba & PyTorch \\ \hline

\end{tabular}
\end{table}

\subsection{Datasets}
The following datasets have been selected for this study:

\begin{enumerate}
\item \textbf{FEMNIST}: This dataset is an extended version of the MNIST dataset, specifically adapted for FL by the LEAF benchmark. It is an image classification dataset comprising 3,550 client nodes and a total of 80,5263 samples (approximately 226.83 samples per client on average). The dataset contains 62 distinct classes (10 digits, 26 lowercase letters, and 26 uppercase letters) and exhibits a non-i.i.d. distribution. In this study, we utilized 200 clients, with a train/test/validation split of 60\%/20\%/20\% for each client.

\item \textbf{CIFAR10}: This well-known dataset, introduced by the University of Toronto, contains 60,000 color images, distributed across 10 classes (6,000 samples per class). Compared to FEMNIST, CIFAR10 is computationally more demanding. We partitioned the data among 50 clients using the Latent Dirichlet Allocation (LDA) algorithm \footnote{LDA is used here to create data heterogeneity}. The train/test/validation split employed in this study is 66.7\%/16.67\%/16.67\%, aligned with splits commonly used in the literature to ensure comparability.
\end{enumerate}

\subsection{Common Hyper-parameters of FL frameworks}
Table \ref{table3} presents some of the common hyperparameters (HPs) used in the FL environment. It is important to note that the names of HPs may vary slightly depending on the specific FL package being used. Further, there are other less common HPs that are not included in this table, as these can vary based on the type of FL architecture (e.g., vertical FL), the objective function, the optimization method, regularization strategies, and so forth.
Although the FL framework itself includes several HPs, such as the number of communication rounds and the federated optimization method, which play a key role in overall system efficiency \cite{17}, determining the optimal global and communication management HPs is a complex task. This challenge is further compounded by the substantial matrix operations required, which depend on the target objective function and the vast number of HP combinations \citeN{3,5}. Although this work mainly focuses on applying conventional HPO tools to the DFL paradigm, local HPs have been selected for optimization in this study. Global and communication management HPs are unique to the FL setting and are generally not supported by most available HPO tools.

\begin{table}[htbp]
\centering
\resizebox{\textwidth}{!}{%
\begin{tabular}{|p{3.5cm}|p{2cm}|p{2cm}|p{2cm}|p{5.5cm}|}
\hline
\textbf{Parameter} & \textbf{Type} & \textbf{Default} & \textbf{Range} & \textbf{About} \\ \hline
training\_type & Common args & Simulation & Cross-silo & \begin{itemize}
  \item Cross-silo stands for cross-organization. In that case, each local node is a server.
  \item Cross-device is used when the local nodes are resource-constrained devices (e.g., cell phones).
  \item Simulation or standalone mode is when all local clients are simulated within one computational node.
\end{itemize} \\ \hline
random\_seed & Common args & '1234' &  & Random Seed for reproducibility \\ \hline
dataset & Data arg & 'FEMNIT' & 'FEMNIST', 'CIFAR 10' & Two benchmarks for FL \\ \hline
partition\_method & Data arg & 'hetero', '0.5' & [0 – 1] & Stands for how to portion the dataset: determines the measure of heterogeneity or homogeneity \\ \hline
model & Model args & 'CNN-convnet2' & CNN, RNN, GAN, etc. & Local and global Model Zoo \footnote{https://modelzoo.co} \\ \hline
federated\_optimizer & Train args & FedAvg & FEDAVG, FEDOPT, FEDNOVA, FEDPROX, FEDGAN, FEDGKT, FEDNAS, FEDSEG & The aggregation method in the global server (or the type of aggregation in decentralized FL) \\ \hline
client\_num & Train args & 200 for 'FEMNIST', 5 for 'CIFAR10' & [1,5000] & Number of total clients participating \\ \hline
Client\_per\_round & Train args & 2 & [2-client\_num] & Number of clients participating in each communication round \\ \hline
comm\_round & Train args & Stop Criteria \footnote{This HP is differing for different FL tools, datasets, and HPO technologies} & Stop Criteria & Number of communication rounds \\ \hline
epochs & Train args & 1 & [1-Stop Criteria] & Number of training epochs in clients \\ \hline
weight\_decay & Train args & 1E-3 & [10\^-3 – 10\^3] & Weight Decay in client models \\ \hline
batch\_size & Train args & 16 & 2,4,8,16,32,...,2n & Batch size of the dataset participating in each training round \\ \hline
client\_optimizer & Train args & 'SGD' & SGD is the most famous one & NN Optimization approach for client nodes \\ \hline
learning\_rate & Train args & 0.03 & [10\^-2 – 10\^4] & Learning rate for training DNNs inside client nodes \\ \hline
using\_gpu & device\_args & 1 & False/True & Determines if GPU is allowed or not \\ \hline
worker\_num & device\_args & 2 & 0-2 & Number of CPU/GPU workers \\ \hline
\end{tabular}%
}
\caption{Overview of the most popular open-source Auto-ML toolkits}
\label{table3}
\end{table}

\subsection{Implementation Details}
In this study, the primary intention is to deploy efficient lightweight HPO tools for FL environments. Our results have been reported by running HPO on both local and global models. Besides the challenges of interaction between different packages, several criteria should be met when selecting FL Libraries and Auto-DL tools. FL tools are required to be adaptable, easy to deploy, and support a developer mode. Further, they should contain both a "standalone" version and a "distributed" version. In the "distributed" version, a real-world cluster node represents a client, while in the "standalone" mode, clients are simulated within a single node. Although resource allocation for each node can be predefined in some HPO libraries, the developments in this study are conducted in the "standalone" simulation mode due to existing resource limitations in the FL setting. 
Ergo, in this work, PySyft 0.8 has been chosen as the FL platform as it satisfies the requirements for an FL library. It offers a modifiable HPO module, which can be deployed with personalized HPO functions. By adding a low-fidelity add-on, a balance between computational time and accuracy can be achieved. Further, the PySyft package is one of the rapidly evolving research-based FL libraries that supports single-machine simulations. This platform promotes the integration and development of different algorithms, making it a suitable candidate for this study.

Deploying HPO technologies for FL platforms also requires meeting certain preconditions. Firstly, these tools should be fast-paced. Conducting HPO for classical ML approaches typically requires significantly more time than one learning round, as many HP configurations must be tested. Since the search space in the FL setting is several times larger than in a general DL problem, the speed of the HPO algorithm becomes even more crucial. Some HPO tools include a built-in low-fidelity module for time-sensitive and real-time tasks where time is prioritized over accuracy. Another essential condition is that the HPO tool should be lightweight. Given the number of participating clients, the number of transferred HPs between clients and the server in each communication round, and the learning parameters within each local node, FL systems are already resource-intensive. Therefore, being lightweight is a critical feature for an HPO tool, particularly when resources are limited. 

Further, the selected HPO tool must support DNN structures and image-based datasets, as this work focuses on a DFL environment with image-based datasets. RayTune and Optuna have been selected as HPO technologies for the FL environment. Optuna is one of the most lightweight HPO tools for DL and is relatively fast. These characteristics, along with the fact that it is a PyTorch package, make it an excellent candidate for the PySyft framework. Conversely, RayTune, known for its fast tuning and low computational overhead when working with large datasets, has been chosen for the FL setting due to its lightweight HPO methodology.

All experiments were executed on a server containing two NVIDIA RTX A6000 GPUs, each with 48 GB of memory and 59 cores. The CUDA version is 11.3, and the experiments were conducted using PyTorch 1.10.1.

The aggregation algorithm used in all experiments in this study is "FedAvg," which is the most widespread server aggregator in the FL literature. It is noteworthy that "FedAvg" employs the Stochastic Gradient Descent (SGD) algorithm to collect local model weights trained on non-i.i.d. datasets \cite{4}. The mean validation loss and average test accuracy have been selected as the objective function for HPO in this study.

\section{Experimental Result} \label{5}
In this paper, we conduct HPO on the Learning Rate (LR), the number of local epochs, and the weight decay of local DNNs inside each client. The range of LR and other HPs can be found in Table \ref{Range_HP}. The evaluation frequency has been set to every five communication rounds, and the mean validation loss and accuracy over these repetitive results have been calculated. Random Search has been selected as the baseline HPO method and compared with both HPO toolkits. Both HPO toolkits have been working in parallel to find the near-optimal HP set in the FL platform using a step-wise adaptive HPO mechanism. The results demonstrate that this step-wise mechanism, when integrated with HPO toolkits, is effective for finding more optimum configurations of HPs among all other HP combinations in the same search space.

\begin{figure}[h]
  \centering
  \begin{subfigure}[b]{0.49\textwidth}
  \label{fig:FedML+FEMNIST}
  \includegraphics[width=\linewidth]{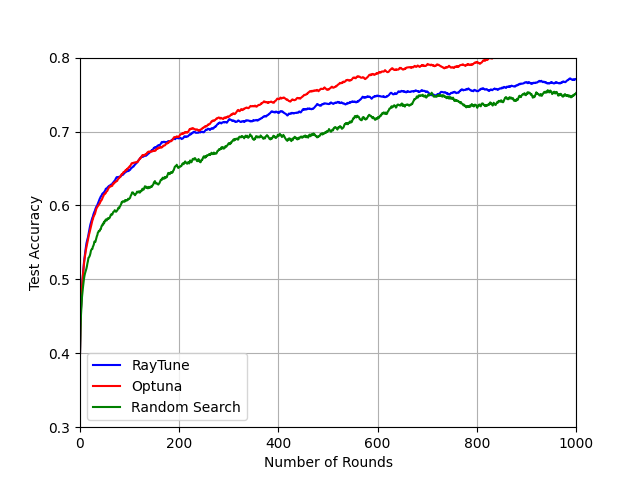}
  \caption{%Comparison of HPO Tools in DFL experimented on FEMINIST data set with 200 client nodes.
  }
   \end{subfigure}
   \hfill
    \begin{subfigure}[b]{0.49\textwidth}
  \centering
  \label{fig:FedML+CIFAR10}
  \includegraphics[width=\linewidth]{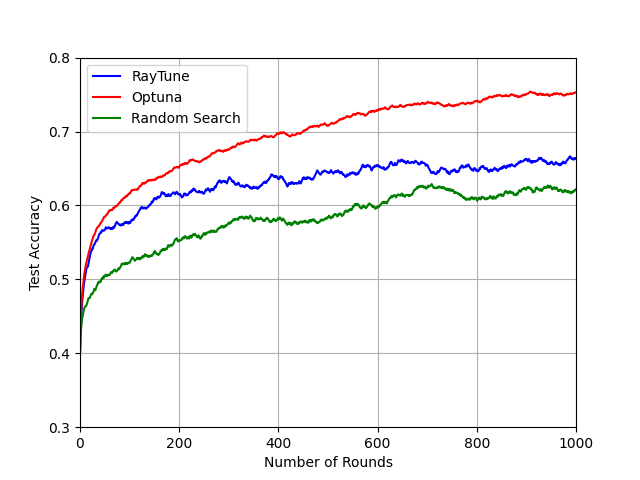}
  \caption{%Comparison of HPO Tools in DFL experimented on CIFAR10 data set for 200 client nodes.
  }
\end{subfigure}
  \begin{subfigure}[b]{0.49\textwidth}
  \label{fig:FedML1} 
  \includegraphics[width=\linewidth]{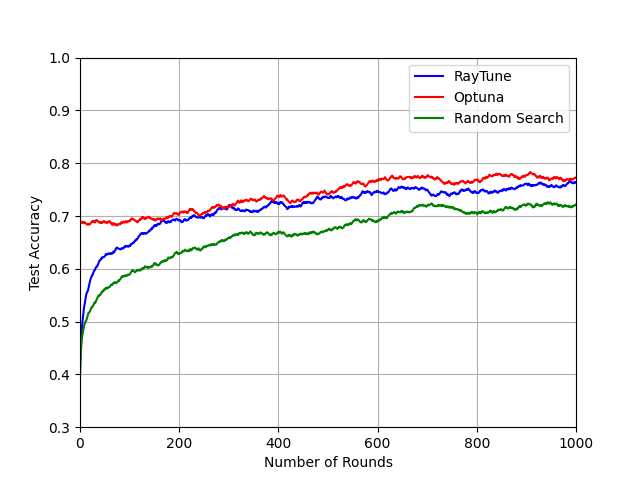}
  \caption{%Comparison of HPO Tools in DFL experimented on FEMINIST data set with 20 clients.
  }
   \end{subfigure}
   \hfill
    \begin{subfigure}[b]{0.49\textwidth}
  \centering
  \label{fig:FedML2}
  \includegraphics[width=\linewidth]{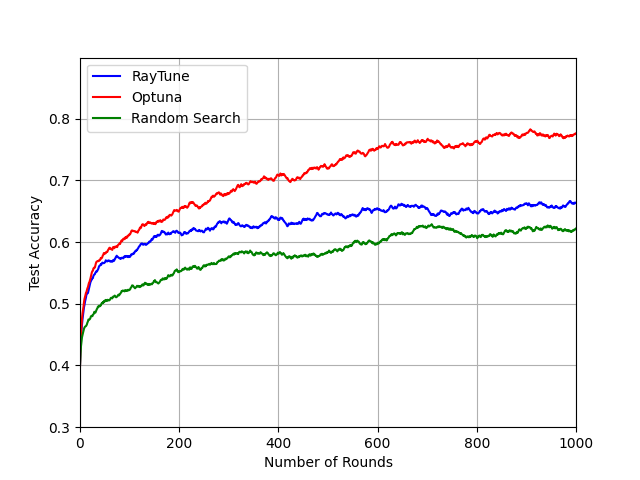}
  \caption{%Comparison of HPO Tools in DFL experimented on CIFAR10 data set with 20 clients.
  }
\end{subfigure}
 \caption{The test accuracy of Random Search, Optuna, and RayTune HPO toolkits for both FEMNIST and CIFAR10 non-i.i.d datasets for 1000 communication rounds on \textbf{(c),(d)} small-scale(20 clients), and \textbf{(a),(b)} large-scale(200-clients) FL setting.}
  \label{fig:both_images}
\end{figure}

The results after applying RayTune and Optuna in DFL for FEMNIST and CIFAR10 are presented in Table \ref{FLAML_FEMNIST} and Table \ref{FLAML_CIFAR10}, respectively. Figure \ref{fig:both_images} illustrates the test accuracy across 1000 communication rounds for Optuna and RayTune compared to Random Search. The results demonstrated that both HPO toolkits successfully identified better HP sets for the DFL platform on the two non-i.i.d. datasets. Optuna achieves higher accuracy in both cases; however, the average search time was lower when RayTune was employed. RayTune employs a parallel search mechanism in distributed training, making it a reliable and efficient HPO tool for FL environments. On the other hand, Optuna was able to reach approximately 81\% accuracy for the FEMNIST dataset despite its slower search speed. 
Optuna's architecture is constructed with an event loop that manages the messaging paradigm in distributed systems. The messages are controlled by the FL trusted server to ensure privacy constraints. Coordinating message exchange with the trusted server in FL poses several challenges in different FL scenarios, such as synchronous versus asynchronous FL. It also depends on the discrepancy between the waiting time for the next communication round and the search time of the Optuna HP sampler. In such cases, the use of a step-wise adaptive HPO mechanism could be advantageous, as it allows Optuna to receive feedback from the group of clients after the weights have been aggregated by the global server.

\begin{table}[]
\centering
\caption{Test results of applying Random Search (before) and applying Optuna and Raytuna as an HPO tool (after) for  Pysyft  on the  FEMINIST.} 
\label{FLAML_FEMNIST}
\begin{tabular}{|l|l|l|l|l|}
\hline
\textbf{FL library + Auto-ML tool} & \textbf{Accuracy} & \textbf{Weight Decay}  & \textbf{Epochs} & \textbf{LR} \\ \hline
Pysyft (default) / Random Search        & 0.7304           & 0.01                                   & 6         & 0.001       \\ \hline
Pysyft + Optuna        & 0.8143            & 0.0001                                            & 2   & 0.01       \\ \hline
Pysyft + Raytune    & 0.7793           & 0.00001                                            & 5   & 0.001         \\ \hline
\end{tabular}
\end{table}

\begin{table}[]
\centering
\caption{Test results of applying Random Search (before) and applying Optuna and Raytuna as an HPO tool (after) for  Pysyft  on CIFAR10.} 
\label{FLAML_CIFAR10}
\begin{tabular}{|l|l|l|l|l|l|}
\hline
\textbf{FL library + Auto-ML tool} & \textbf{Accuracy} & \textbf{Weight Decay} & \textbf{Epochs}  & \textbf{LR}  \\ \hline
Pysyft + RS           & 0.6087            & 0.01                                        & 3      & 0.001                       \\ \hline
Pysyft + Optuna    & 0.7323            & 0.00001                                       & 7     & 0.01                       \\ \hline
Pysyft + Raytune    & 0.6809           & 0.001                                        & 5      & 0.01                       \\ \hline
\end{tabular}
\end{table}

\begin{table}[]
\centering
\caption{Range of Hyper-parameters and their step in low-fidelity mode.} 
\label{Range_HP}
\begin{tabular}{|l|l|l|}
\hline
\textbf{Hyper-parameter} & \textbf{range} & \textbf{Step}  \\ \hline
Weight Decay           & $ [1e^{-5} - 1e^{-1}] $            & $e$                                         \\ \hline
Local training epochs    & [0-10]            & 1                                                     \\ \hline
Learning Rate    &  $ [10^{-5} - 10^{-1}] $     & 10                                                             \\ \hline
batch Size    &  $[2^4 - 2^8]$        & 2                                                            \\ \hline
Drop out    &  $[0.1-0-5]$        & 0.2                                                           \\ \hline
\end{tabular}
\end{table}

% \begin{figure}[h]
%   \centering
%   \begin{subfigure}[b]{0.55\textwidth}
%   \label{fig:FedML1} 
%   \includegraphics[width=\linewidth]{samples/Figure/FEMNIST-4.png}
%   \caption{Comparison of HPO Tools in DFL experimented on FEMINIST data set with 20 clients.}
%    \end{subfigure}
%    \hfill
%     \begin{subfigure}[b]{0.55\textwidth}
%   \centering
%   \label{fig:FedML2}
%   \includegraphics[width=\linewidth]{samples/Figure/CIFAR10-4.png}
%   \caption{Comparison of HPO Tools in DFL experimented on CIFAR10 data set with 20 clients.}
% \end{subfigure}
%  \caption{The test accuracy of Random Search, Optuna, and RayTune HPO toolkits for both FEMNIST and CIFAR10 non-i.i.d datasets for 1000 communication rounds on a small FL environment.}
%   \label{fig:both_images}
% \end{figure}

The effect of scalability has also been explored in this work. Figures 4 and 5 illustrate the comparison between the aforementioned Auto-FL tools for a small-scale FL environment with only 20 clients. The results illustrate that both Optuna and RayTune continue to outperform Random Search. Nonetheless, there are a few differences, mainly due to the lack of diversity and quantity of datasets in a small number of clients compared to a large cohort of nodes. First off, the slope of the accuracy charts is less steep in small-scale FL platforms compared to large-scale environments. This is particularly evident in CIFAR10, where the limited data distribution could slow the learning process. Secondly, these results indicate that Optuna performs optimally in large-scale FL platforms with a greater number of nodes. Moreover, the CIFAR10 dataset results show no significant differences when comparing different scales (20 or 200 clients).  

 The results demonstrate the efficacy of integrating Auto-ML toolkits into FL for model adaptation. As discussed earlier, Auto-FL significantly reduced the time and effort required for hyperparameter tuning. It achieved competitive hyperparameter configurations within a few federated learning rounds, whereas manual tuning typically requires extensive trial and error. Auto-FL also exhibited less training time, with fewer communication rounds completing up to 10\% and 20\% faster compared to Random Search for Optuna and RayTune, respectively. This efficiency can be attributed to the ability of Optuna and RayTune AutoML tools to optimize local training settings and their lightweight nature. Further, the implemented Auto-FL environment demonstrated impressive scalability, maintaining consistent performance in both small-scale and large-scale FL settings. This suggests that Auto-FL toolkits can effectively handle federated learning scenarios with a large and diverse set of clients.

\section{Conclusion and Future works}\label{6}
As the demand for efficient FL systems increases, many researchers are seeking to integrate AutoML approaches with FL environments. This work addressed the HPO-FL problem by exploring and designing the integration of well-known HPO toolkits within the FL setting. Several AutoML tools have been discussed in this paper, and two lightweight AutoML tools have been applied to the Deep Federated Learning framework to find the optimal HP sets for the local HPs within client nodes. The experiments in this study demonstrate the effectiveness of utilizing existing well-known HPO tools on two benchmark datasets, CIFAR10 (non-i.i.d.) and FEMNIST. Further, this work introduces a step-wise feedback mechanism for deploying HPO toolkits in FL platforms, allowing for an accelerated, more synchronized tuning process. In summary, our experiments highlight the advantages of using AutoML toolkits (Optuna and RayTune) within the Pysyft Federated Learning framework. Auto-FL reduced the efforts required for HP tuning, improved efficiency, and demonstrated scalability across two distinct datasets. These findings underscore the potential of integrating AutoML tools into FL as a powerful approach for automating and optimizing FL processes across various domains. 

While this work represents a first step towards optimizing the FL setting with AutoML tools, future research could focus on integrating additional AutoML techniques and evaluating Auto-FL in real-world federated learning applications. A few future directions for applying AutoML methods in FL are summarized below. \\ 

\begin{itemize}
\item \textbf{Neural Architecture Search (NAS)}: Since NAS has demonstrated excellent results in automating the design of DNN architectures and tuning their HP sets, FL researchers are eager to integrate it into FL environments. Although the number of studies in this area is increasing, one must consider the high computational cost associated with existing NAS approaches. To date, running NAS is only feasible for client nodes equipped with powerful GPUs. Nevertheless, researchers could adopt lightweight NAS techniques for resource-constrained devices by constraining the search space and removing large DNN architectures. 

\item \textbf{Global HPs}: As the primary goal of this study was to explore the integration of HPO tools for classic DL tasks in FL environments, HPs specific to the FL structure were not examined. Global HPs, such as the $\lambda$ factor and optimization functions, are essential aspects of HPO-DFL that should be investigated further. It is noteworthy  that global and communication management HPs can directly influence many objectives of FL systems, including fairness, robustness, personalization and privacy, and thus require more precautions when tuned. 
\item \textbf{Low-fidelity HPO approaches}: HPO in FL frameworks often requires significant computational resources and time due to the distributed and heterogeneous nature of the data. To mitigate this, low-fidelity HPO methods offer a promising direction. For instance, researchers from \textit{FedScope} have introduced a low-fidelity feature for the Auto-FL task, which improves time efficiency by reducing the hyper-parameter search space. There is considerable potential for developing novel low-fidelity HPO techniques, such as employing small-scale DNNs on each client or leveraging group knowledge transfer techniques to reduce computational overhead \cite{1}. Further strategies could include: 1) reducing the participation rate by allowing inactive clients to drop out of communication rounds without impacting global convergence, 2) performing partial local HPO by sampling smaller batches of data from the local distribution in each training round, which balances accuracy and resource efficiency, 3) employing multi-fidelity optimization techniques that dynamically allocate more resources to clients or configurations with higher performance potential, and 4) incorporating clustering techniques to share optimized hyper-parameters across similar clients, accelerating the HPO process in future rounds.
\end{itemize}

\section*{Acknowledgment}

This work is  partly based upon the work supported by the National Center for Transportation Cybersecurity and Resiliency (TraCR) (a U.S. Department of Transportation National University Transportation Center) headquartered at Clemson University, Clemson, South Carolina, USA. Any opinions, findings, conclusions, and recommendations expressed in this material are those of the author(s) and do not necessarily reflect the views of TraCR,  and the U.S. Government assumes no liability for the contents or use thereof.

\bibliographystyle{ACM-Reference-Format}
\bibliography{Auto_FL_REF}

\appendix

\end{document}